\documentclass[sigconf]{acmart}

\PassOptionsToPackage{prologue,table,xcdraw}{xcolor}

\usepackage{amsmath,graphicx}
\usepackage{multirow}
\usepackage{graphicx}
\usepackage{lscape}
\usepackage[table,xcdraw]{xcolor}
\usepackage{diagbox}

\AtBeginDocument{%
  \providecommand\BibTeX{{%
    \normalfont B\kern-0.5em{\scshape i\kern-0.25em b}\kern-0.8em\TeX}}}

\copyrightyear{2022} 
\acmYear{2022} 
\setcopyright{acmcopyright}\acmConference[WWW '22 Companion]{Companion Proceedings of the Web Conference 2022}{April 25--29, 2022}{Virtual Event, Lyon, France}
\acmBooktitle{Companion Proceedings of the Web Conference 2022 (WWW '22 Companion), April 25--29, 2022, Virtual Event, Lyon, France}
\acmPrice{15.00}
\acmDOI{10.1145/3487553.3524721}
\acmISBN{978-1-4503-9130-6/22/04}

\begin{document}

\title{CCGG: A Deep Autoregressive Model for Class-Conditional Graph Generation}


\author{Yassaman Ommi}
\authornotemark[1]
\affiliation{%
  \institution{Amirkabir University of Technology}
  \city{Tehran}
  \country{Iran}
}
\email{yassi.ommi@aut.ac.ir}

\author{Matin Yousefabadi}
\authornote{Both authors contributed equally to this research.}
\affiliation{%
  \institution{Sharif University of Technology}
  \city{Tehran}
  \country{Iran}}
\email{mtnusf97@gmail.com}

\author{Faezeh Faez}
\affiliation{%
  \institution{Sharif University of Technology}
  \city{Tehran}
  \country{Iran}}
\email{faezeh.faez@gmail.com}

\author{Amirmojtaba Sabour}
\affiliation{%
  \institution{Sharif University of Technology}
  \city{Tehran}
  \country{Iran}}
  \email{amsabour79@gmail.com}

\author{Mahdieh Soleymani Baghshah}
\affiliation{%
  \institution{Sharif University of Technology}
  \city{Tehran}
  \country{Iran}}
\email{soleymani@sharif.edu}

\author{Hamid R. Rabiee}
\affiliation{%
  \institution{Sharif University of Technology}
  \city{Tehran}
  \country{Iran}}
\email{rabiee@sharif.edu}

\renewcommand{\shortauthors}{Yassaman Ommi et al.}

\begin{abstract}
Graph data structures are fundamental for studying connected entities. With an increase in the number of applications where data is represented as graphs, the problem of graph generation has recently become a hot topic. However, despite its significance, conditional graph generation that creates graphs with desired features is relatively less explored in previous studies. This paper addresses the problem of class-conditional graph generation that uses class labels as generation constraints by introducing the Class Conditioned Graph Generator (CCGG). We built CCGG by injecting the class information as an additional input into a graph generator model and including a classification loss in its total loss along with a gradient passing trick. Our experiments show that CCGG outperforms existing conditional graph generation methods on various datasets. It also manages to maintain the quality of the generated graphs in terms of distribution-based evaluation metrics.
\end{abstract}

%


\maketitle

\section{Introduction}
Graphs are ubiquitous in many areas of science and engineering. Graph generation is one of the essential research lines in graph studies, which dates back to several decades ago \cite{erdHos1960evolution}. In recent years, with the increasing popularity of deep generative models, graph generation research has again become a hot topic, extending a various number of deep learning frameworks, with applications ranging from drug discovery to social network analysis \cite{li2018multi} \cite{faez2021deep}.
However, in generating new samples, one may want to impose certain conditions and characteristics on their outputs to benefit various domains. This condition can be based on class labels, text data, images, etc. While this problem is well-investigated in other domains, it has been relatively less approached in the field of graph generation due to its recent emergence. Consequently, the subproblem of class-conditional graph generation, which gives the generative model the ability to generate graphs of desired classes, had not been addressed despite its importance and applications.

\begin{figure*}[!t]
\centering
\includegraphics[scale=0.15]{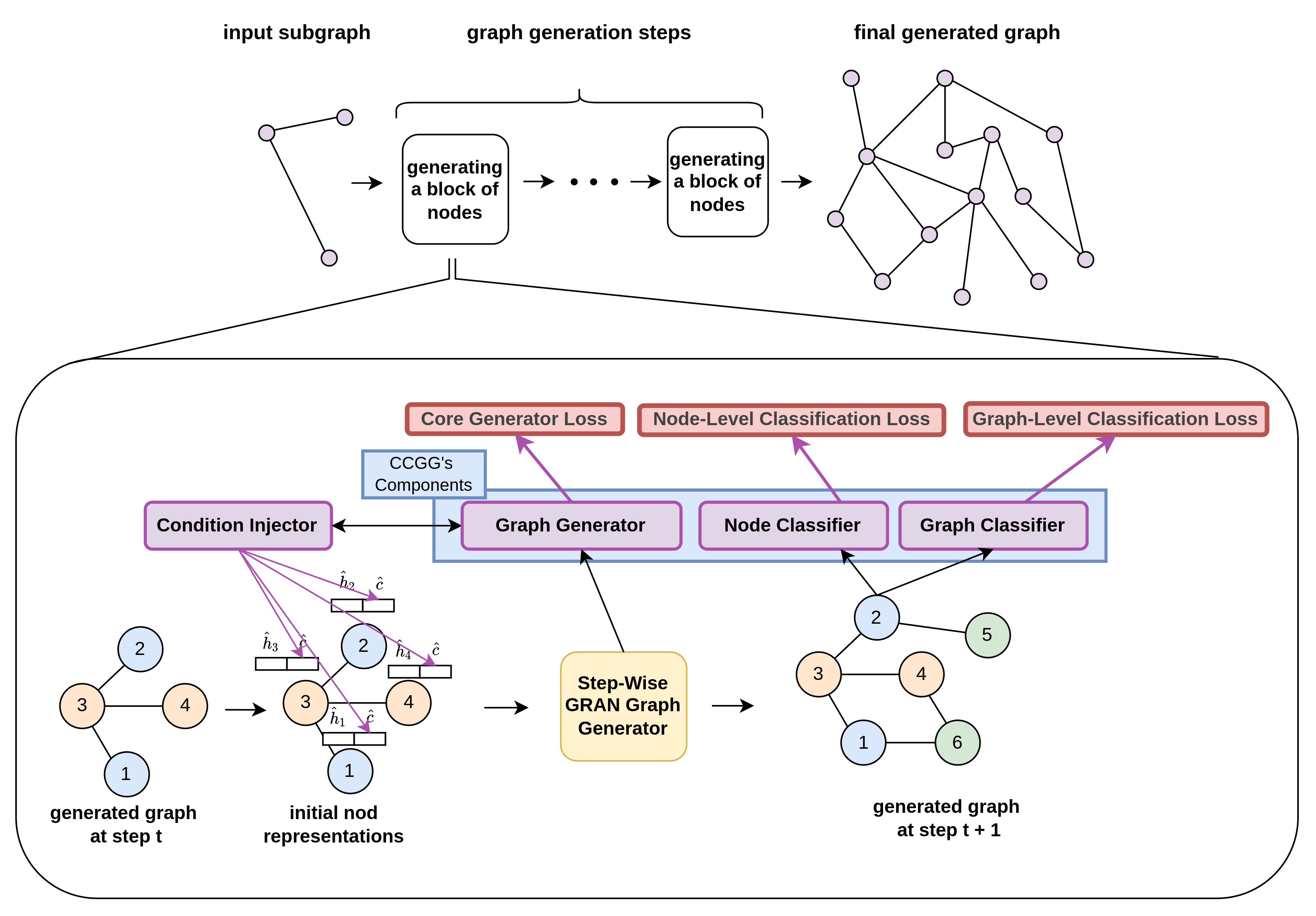}
\caption{Overview of the proposed CCGG model. (top figure) Multiple steps of generating node blocks generate the final desired graph from an input subgraph. (bottom figure)  Nodes with the same color are generated in the same step. The graph generator component generates new nodes and associated edges utilizing the condition injector in each graph generation step. Then using the node classifier and graph classifier, the node-level classification loss and the graph-level classification loss are added to our final loss function.}
\label{fig_sim}
\end{figure*}

\indent In this paper, we study the novel problem of class-conditioned graph generation, whose goal is to learn and generate graph structures given the class information. In this regard, we propose CCGG, based on a deep generative model of graphs. Specifically, we learn a likelihood over graph edges via an autoregressive generative model of graphs, i.e., GRAN \cite{liao2019efficient} built upon graph recurrent attention networks. At the same time, we inject the graph class information into the generation process and incline the model to generate graphs of desired classes by introducing a novel graph classification loss and the graph generation loss function. Unlike GRAN, which is a non-conditional graph generator, our method can conditionally generate graphs of specified categories while maintaining the time efficiency and quality of generated graphs. This problem is relatively unexplored despite its importance and applications.
Our proposed model consists of four main parts. The core generator component generates a  graph in a step-by-step manner. Then, we use a graph classifier we have trained using the GraphSAGE framework \cite{hamilton2017inductive}, to incline the model towards generating graphs of our desired classes. Moreover, our node-level classifier makes CCGG capable of better learning graphs distribution in each class. Lastly, we include two new elements in our loss functions to capture the requirements of the classification task. The overview of our model is depicted in Fig \ref{fig_sim}.

\indent We summarize the key contributions of our work as follows:
\begin{itemize}
    \item  We address the novel problem of class-conditional generation, which has not been directly addressed in the past, by injecting the class indicator into the primary model.
\item Unlike the previously proposed generator, we utilize GNNs to better capture the input graphs’ topology, supporting the generation of more realistic graphs.
\item We have used node classifiers, which were relatively less employed by the previous studies, improving the quality of the generated samples. 
\item We have defined a novel loss function that can successfully take the model’s different components into account while also facilitating the task of generating larger graphs from smaller initial graphs.
\end{itemize}


\section{Related Work}

\subsection{Graph Generators}
Recent advances in deep learning, the limitations of classic graph generation methods, and the field's broad application have resulted in the development of deep graph generators over the past few years. The goal of graph generators is to either implicitly or explicitly learn a distribution based on a train set of observed graphs and to be able to sample from the said distribution to obtain new graphs with similar properties to the train set. In general, these methods can be categorized according to their framework into autoregressive, VAE-based, GAN-based, RL-based, and flow-based models, as introduced in \cite{faez2021deep}. For instance, \cite{assouel2018defactor} proposes an autoregressive graph generator, utilizing the LSTM model, while \cite{simonovsky2018graphvae, guo2020interpretable} are latent space-based models, extending the VAE framework. Moreover, there exist methods that use generative adversarial nets (GANs), like \cite{gamage2020multi, bojchevski2018netgan} that operate on random walks to create realistic graphs, or \cite{de2018molgan} that takes the entire graph as input.  

\subsection{Conditional Generators}
Conditional sample generation is a fundamental problem for generative methods with many applications. It is an advantageous feature for generative models, as it provides the means of inducing desired characteristics in the generated outputs. In this regard, this problem has been widely explored in various domains such as image and text. Many text and image generators are conditioned on multiple input types. For instance, \cite{Wang2018SentiGANGS} proposed SentiGAN, a GAN-based model to generate sentimental text under the condition of specific emotional class labels, while \cite{wang2019topicguided} extends the VAE framework to build a text generator conditioned on a given topic label. Moreover, \cite{kang2020contragan, Yan2016attribute2image, 8302049} propose deep image generators, applying specific attributes and conditions to their generated output. On the other hand, there exist generators that are conditioned on a sequence of inputs rather than a single label \cite{DBLP:journals/corr/abs-1909-03409}. For example, \cite{li2019storygan} visualizes a story by generating a sequence of images given a multi-sentence paragraph, \cite{cheng2020rifegan} creates realistic images based on a provided description, and \cite{zhang2021text} proposes a method to manipulate an image based on a textual instruction. However, there is still much work for conditional graph generation compared to similar studies in other domains, such as text and image. \\
This issue is also a concern with graph data structures, where constructing graphs with given domain-specific conditions can be of great significance, such as particular molecular optimization for drug discovery \cite{jin2019learning}. Considering this, GraphVAE \cite{simonovsky2018graphvae} presents a conditional setting on their model, by utilizing a molecule decoder, for the specified task of molecular graph generation. Though they still can't guarantee the generated molecules' semantic (chemical) validity. Recently, AGE \cite{fan2020attention} has proposed a deep generative model with attention, which can be conditioned on an input graph and outputs a transformed version of it, which can be analogous to its evolution. Existing graph generation methods have approached this problem with varying procedures like directly injecting validity constraints \cite{ma2018constrained}, or latent space optimization \cite{liu2018constrained}. Furthermore, \cite{yang2019conditional} proposes a GAN-based conditional graph generator that contacts a condition vector to the nodes’ latent representations to make their model able to generate graphs conditionally. However, recent graph generation studies have not directly studied the class-conditional generation problem. \\

\section{Problem Formulation}
We focus on the novel problem of class-conditional graph generation. Suppose we are given a set of graphs $G=\{G_1, G_2, ..., G_n\}$ with the set of graph classes $C=\{C_1, C_2, ..., C_m\}$, where $G_i = (V_i, E_i) \in C_j$ denotes a graph of the class $C_j$ with the set of nodes and edges $V_i$ and $E_i$, respectively. Our goal is to train a graph generative model on a dataset of graphs to generate new graph structures given the desired class label at inference time.

\section{Method}
\label{sec:method}
This section presents our CCGG model, a deep autoregressive model for the class-conditional graph generation. First, we introduce the GRAN model that we employ as a powerful GNN-based graph generator in our model. Then, we present the proposed conditional graph generation model.
\subsection{Background: GRAN model}
\label{IIA}
Many advances have been made to introduce deep models for generating graphs regardless of their structures by estimating their distribution in recent years. Among recently proposed methods for the graph generation problem, we have utilized GRAN \cite{liao2019efficient} for our study. Compared to other previous methods like GraphRNN \cite{you2018graphrnn}, GRAN uses Graph Neural Networks with an attention mechanism, which enables it to reach state-of-the-art results while preserving scalability to much larger graphs. 

For an undirected graph $G=(V, E)$, where $V$ is the graph's node-set, and $E$ represents its edge set, $A^{\pi}$ can be considered as the adjacency matrix of the graph, where $\pi$ is a specific node ordering of the graph. Therefore, $L^{\pi}$ can be denoted as the lower triangular part of $A^{\pi}$ and we have $A^{\pi}=L^{\pi}+L^{\pi \top}$. GRAN then generates the graphs by utilizing a step-by-step approach where at each step $B$ rows of $L^{\pi}$ is generated. Each block in the $t^{t h}$ step, is represented as $L_{\boldsymbol{b}_{t}}^{\pi}$, and is a sequence of vectors $L_{i}^{\pi} \in {R}^{1 \times|V|}$, where $i \in  \boldsymbol{b}_{\boldsymbol{t}}=\{B(t-1)+1, \ldots, B t\}$. Thus, at each generation step, GRAN tries to optimize the following likelihood to generate one block of nodes and its associated edges: 
 \begin{equation}
p\left(L^{\pi}\right)=\prod_{t=1}^{T} p\left(L_{\boldsymbol{b}_{t}}^{\pi} \mid L_{\boldsymbol{b}_{1}}^{\pi}, \cdots, L_{\boldsymbol{b}_{t-1}}^{\pi}\right).
\end{equation}

GRAN then assumes a representation for each node at each step and updates them using the following GNN with attentive messages:

\begin{equation}
m_{i j}^{r}=f\left(h_{i}^{r}-h_{j}^{r}\right)
\end{equation}
\begin{equation}
\tilde{h}_{i}^{r}=\left[h_{i}^{r}, x_{i}\right]
\end{equation}
\begin{equation}
a_{i j}^{r}=\operatorname{Sigmoid}\left(g\left(\tilde{h}_{i}^{r}-\tilde{h}_{j}^{r}\right)\right)
\end{equation}
\begin{equation}
a_{i j}^{r}=\operatorname{Sigmoid}\left(g\left(\tilde{h}_{i}^{r}-\tilde{h}_{j}^{r}\right)\right)
\end{equation}
\begin{equation}
h_{i}^{r+1}=\operatorname{GRU}\left(h_{i}^{r}, \sum_{j \in \mathcal{N}(i)} a_{i j}^{r} m_{i j}^{r}\right)
\end{equation}

\noindent where $h_{i}^{r}$ is the representation for node $i$ after round $r$ , $m_{ij}^{r}$ is the message vector from node $i$ to $j$, $x_{i}$ indicates whether node $i$ is in the previously generated nodes or the newly added ones, and $a_{ij}^{r}$ is an attention weight associated with edg $e(i, j)$. Here, $f$ and $g$ are 2-layer MLP with ReLU nonlinearity.

Finally, after R rounds of updating the node representations, the edges connecting the nodes of the $t^{th}$ block to other nodes are generated using a mixture of Bernoulli distributions. The parameters of the distributions are obtained via two MLPs with ReLU nonlinearities, using the final node representations derived from the last step. 

\begin{equation}
\alpha_{1}, \ldots, \alpha_{K}=\operatorname{Softmax}\left(\sum_{i \in b_{t}, 1 \leq j \leq i} \operatorname{MLP}_{\alpha}\left(h_{i}^{R}-h_{j}^{R}\right)\right),
\end{equation}
\begin{equation}
\theta_{1, i, j}, \ldots, \theta_{K, i, j}=\text { Sigmoid }\left(\operatorname{MLP}_{\theta}\left(h_{i}^{R}-h_{j}^{R}\right)\right).
\end{equation}
Here, $\alpha_{i}$ and $\theta_{k, i, j}$ are the parameters associated with the mixture of Bernoulli distributions. Therefore, the probability of generating the block $L_{\boldsymbol{b}_{t}}^{\pi}$ can be written as:

\begin{equation}
p\left(L_{b_{t}}^{\pi} \mid L_{b_{1}}^{\pi}, \ldots, L_{b_{t-1}}^{\pi}\right)=\sum_{k=1}^{K} \alpha_{k} \prod_{i \in b_{t}} \prod_{1 \leq j \leq i} \theta_{k, i, j}.
\end{equation}

\subsection{Class Conditional Graph Generation}

This section presents our CCGG model, a deep autoregressive model for the class-conditional graph generation. The method adopts the GRAN model \cite{liao2019efficient} as the core generation strategy. Moreover, CCGG consists of three main components. As the first component, the core generator component uses a condition injector that enters the class labels of the graphs as conditions into the training procedure. Secondly, we employ a graph classifier to conduct the graph generator to produce graphs holding the desired condition (i.e., class label). We adopt a node classifier as the third component, enhancing the model's generative process. Moreover, we introduce our customized loss function that combines the results of all components to enable the CCGG in capturing the class-based structures of graphs regardless of their size. Finally, we will elaborate on the approach we employed to deal with the gradient flow problem we faced during the training phase.

\subsubsection{The Core Generator Component}
\label{node_label}
This component is responsible for generating new edges and nodes of the graphs. Here, we inject the class labels of the graph into the node representations such that CCGG can approximate class distributions, and the dataset's general distribution. To do this, at the $t^{th}$ step of generation, we first obtain an initial node representation for the previously generated nodes by concatenating the outputs of two linear mappings:

\begin{equation}
\hat{h}_{\boldsymbol{b}_{i}}=W^{h} L_{\boldsymbol{b}_{i}}^{\pi}+b^{h}, \quad \forall i<t ,
\end{equation}
\begin{equation}
\hat{c}=W^{c} c+b^{c} ,
\end{equation}
\begin{equation}
h_{b_{i}}^{0}=\left(\hat{h}_{\boldsymbol{b}_{i}}, \hat{c}\right) .
\end{equation}
Where $L_{\boldsymbol{b}_{i}}^{\pi}$ is a block of vectors sized $BN$, representing the elements of $L^{\pi}$ in the rows $[B(i-1)+1, \ldots, B i]$, and $N$ is the maximum allowed number of graph nodes. Thus, $\hat{h}_{b_{i}}$ is the pre-initial representation for the $i^{th}$ block nodes, calculated via a linear mapping. Moreover, $c$ is the class-conditional vector of the graph and $\hat{c}$ is obtained by applying a linear mapping on it. These linear mappings are mainly utilized to reduce the size of vectors in large graphs and make the model more flexible in modeling features of each class. Finally, we obtain the initial node representations of $h_{b_{i}}^{0}$, by concatenating $\hat{h}_{\boldsymbol{b}_{i}}$ and $\hat{c}$, which enables the model to generate new blocks with respect to the desired conditions on the graph. Since, $L_{\boldsymbol{b}_{t}}^{\pi}$ is not yet generated for the current block, we set $\hat{h}_{\boldsymbol{b}_{t}}$ = $\boldsymbol{0}$.
Finally, similar to the GRAN approach, the final node representations $h_{i}^{R}$ are obtained by feeding the initial representations to a GNN in $R$ steps, and the edges connecting the nodes are generated using a mixture of Bernoulli distributions. In the test phase, the model generates a graph based on a given class label, which is then fed to the graph classifier.

\subsubsection{The Graph Classifier Component}

To empower CCGG to generate class-conditioned graphs, we added a graph classifier component, with the responsibility of classification during the training phase, to help us calculate the associated graph classification loss. To this end, we have employed GraphSAGE \cite{hamilton2017inductive}, a promising framework for inductive representation learning on large graphs, due to its state-of-the-art results. We trained this graph classifier before our core generation process starts, by attaching a fully connected layer with a sigmoid activation function to the GraphSAGE model. Then, we use this trained model during each step of graph generation by feeding the graph generated so far to it and calculating the classification loss. 

\subsubsection{The Node Classifier Component}

In most graph datasets, nodes have labels indicating their specific features. Therefore, considering these valuable node information can impact the results of models dealing with these datasets. In this regard, we include a node classifier component to increase the accuracy of graph generation in each class. This component helps CCGG better fit the generated graphs' distribution to the original distribution in each class. For this reason, we have appended a two-layer MLP with ReLU nonlinearity as the node classifier, which uses the final node representations explained in the previous section as its input. Moreover, at each step of graph generation, we exploit this component to better capture the class-conditioned characteristics by calculating a node classification loss which will be explained further in the next section.


\begin{table*}[]
\caption{Comparison of CCGG and the baseline model's generated graphs' statistical properties. The Real row indicates the value of the original graphs, while other rows include the absolute values of the differences between the generated and the original graphs. Smaller values of the absolute differences indicate better performance.}
\vspace{-1em}
\begin{center}
\centering
\resizebox{400pt}{!}{%
\begin{tabular}{|c|c|c|c|c|c|c|c|c|c|c|c|}
\hline
\multirow{2}{*}{Dataset} & Properties & \multicolumn{2}{c|}{LCC} & \multicolumn{2}{c|}{TC} & \multicolumn{2}{c|}{CPL} & \multicolumn{2}{c|}{Mean D} & \multicolumn{2}{c|}{GINI} \\ \cline{2-12} 
 & \backslashbox{Models}{Classes} & 0 & 1 & 0 & 1 & 0 & 1 & 0 & 1 & 0 & 1 \\ \hline
\multirow{3}{*}{NCI1} & \textbf{Real} & 25.206 & 36.052 & 0.034 & 0.092 & 4.974 & 6.239 & 2.163 & 2.194 & 0.085 & 0.120 \\ \cline{2-12} 
 & CONDGEN & 9.971 & 16.043 & 15.026 & 27.097 & 2.605 & 3.726 & 0.839 & 0.620 & 0.338 & 0.394 \\ \cline{2-12} 
 & CCGG & \textbf{1.225} & \textbf{1.416} & \textbf{1.873} & \textbf{2.397} & \textbf{0.925} & \textbf{1.647} & \textbf{0.207} & \textbf{0.264} & \textbf{0.022} & \textbf{0.032} \\ \hline
\multirow{3}{*}{PROTEINS} & \textbf{Real} & 47.670 & 22.267 & 34.302 & 17.242 & 5.605 & 3.330 & 3.798 & 3.641 & 0.052 & 0.035 \\ \cline{2-12} 
 & CONDGEN & 19.625 & 6.578 & 155.396 & 38.912 & 3.741 & 1.662 & \textbf{0.149} & 1.520 & 0.479 & 0.255 \\ \cline{2-12} 
 & CCGG & \textbf{3.330} & \textbf{2.347} & \textbf{73.205} & \textbf{11.051} & \textbf{3.009} & \textbf{1.074} & 1.679 & \textbf{0.727} & \textbf{0.056} & \textbf{0.005} \\ \hline
\end{tabular}
}
\label{gini}
\end{center}

\small
\begin{center}
\caption{Graph classification accuracy of each individual class}

\centering
\resizebox{300pt}{!}{%
\begin{tabular}{|c|c|c|c|c|c|c|c|}
\hline
\multirow{2}{*}{Dataset} & \multirow{2}{*}{Classes} & \multicolumn{2}{c|}{GraphSAGE} & \multicolumn{2}{c|}{DiffPool} & \multicolumn{2}{c|}{DGCNN}   \\ \cline{3-8} 
                     &   & CONDGEN  & CCGG              & CONDGEN  & CCGG               & CONDGEN  & CCGG              \\ \hline
\multirow{2}{*}{NCI1} & 0 & 55.32 \% & \textbf{61.00\%}  & \textbf{71.73} \% & 70.37 \%           & \textbf{64.81} \% & 51.61 \%          \\ \cline{2-8} 
                     & 1 & 78.07 \% & \textbf{81.91 \%} & 62.11 \% & \textbf{67. 01 \%} & 69.27 \% & \textbf{74.71 \%} \\ \hline
\multirow{2}{*}{PROTEINS} & 0                        & 74.51 \%  & \textbf{75.16 \%}  & 81.62 \%  & \textbf{88.81 \%} & 71.09 \% & \textbf{71.55 \%} \\ \cline{2-8} 
                     & 1 & \textbf{58.47} \% & 56.23 \%          & 63.13 \% & \textbf{65.24 \%}  & 55.38 \% & \textbf{60.48 \%} \\ \hline
\end{tabular}
}
\label{accuracytable}
\end{center}
\end{table*}



\begin{table}[]
\caption{Graph classification quality measured using AUC (Area Under ROC Curve: Measures how well a parameter or a model can distinguish between right and wrong connections.)}
\vspace{-1em}

\small
\begin{center}
\centering
\resizebox{\columnwidth}{!}{%
\small
\begin{tabular}{|c|c|c|c|c|c|c|}
\hline
\multirow{2}{*}{Dataset} & \multicolumn{2}{c|}{GraphSAGE} & \multicolumn{2}{c|}{DiffPool} & \multicolumn{2}{c|}{DGCNN} \\ \cline{2-7} 
        & CONDGEN & CCGG          & CONDGEN & CCGG          & CONDGEN       & CCGG          \\ \hline
NCI1     & 0.66    & \textbf{0.71} & 0.63    & \textbf{0.65} & \textbf{0.65} & 0.63          \\ \hline
PROTEINS & 0.67    & \textbf{0.69} & 0.72    & \textbf{0.77} & 0.59          & \textbf{0.67} \\ \hline
\end{tabular}
}
\label{auctable}
\end{center}
\end{table}

\subsubsection{Loss Function}

Considering the components mentioned above, the loss function of CCGG consists of three main parts. First, we have the loss of the core generating component, which is the negative log-likelihood of the generated graphs, $\log p(G)=\log \sum_{\pi} p(G, \pi)$, where $\pi$ is a node ordering of the graph. Inspired by GRAN, we consider a family of canonical node orderings to estimate the true log-likelihood due to the factorial number of orders in terms of graph nodes. 
Second, we have the graph-level classification loss denoted as  $\mathcal{L}_{\text {condition }}$, which is calculated as the cross-entropy loss of the generated samples, i.e., $\mathcal{L}_{\text {classifier}}$, multiplied by a power of the discount factor $\gamma$ at each step:

\begin{equation}
\mathcal{L}_{\text {condition }}=\gamma^{|G|-|\hat{G}|} \times \mathcal{L}_{\text {classifier }}.
\label{gamma}
\end{equation}
 $|G|$ and $\hat{|G|}$ represent the number of the original graph's and the input subgraph's nodes, respectively. The idea behind using the discount factor is that, in the generation process, the closer the size of the generated sample to the original graph, the easier it would be for the classifier to classify it. Therefore, as we expect that the classifier performs better on the samples with close nodes to the original ones, their associated classification loss is weighted more.
 
Third, we include the node-level classification loss $\mathcal{L}_{\text{node-label}}$, which is the cross-entropy loss of the generated node label $\mathcal{L}_{\text{node-classifier}}$, calculated the same way as $\mathcal{L}_{\text {condition }}$, multiplied by a power of the reduction factor, exactly like before and with the same intuition.

\begin{equation}
\mathcal{L}_{\text {node-label }}=\gamma^{|G|-|\hat{G}|} \times \mathcal{L}_{\text {node-classifier }}.
\end{equation}
Summing up these losses, the total loss for the CCGG model can be written as:

\begin{equation}\label{eq:total_loss}
\mathcal{L}=\mathcal{L}_{\text {adj }}+\lambda_{1} \mathcal{L}_{\text {condition }}+\lambda_{2} \mathcal{L}_{\text {node-label }}.
\end{equation}
Where $\lambda_{1}$ and $\lambda_{2}$ are the model's hyperparameters, weighting each loss in the training phase. They can be initiated separately for each dataset.

\subsubsection{Solving The Problem in Gradient Flow}

The discrete nature of graphs will cause a gradient flow problem in backpropagating the classifier's loss $\mathcal{L}_{\text {condition }}$. Specifically, to feed adjacency matrices to the classifier, we need to sample the output Bernoulli distributions at each step, which makes the backpropagation intractable. We sidestep this problem by utilizing a categorical reparameterization method, which uses a novel parameterizable Gumbel-Softmax distribution to replace a non-differentiable sample from a categorical distribution with a differentiable sample, a technique proposed in \cite{jang2017categorical}. Following this method, we sample stochastic binary neurons for generating new edges in the training phase, and for the evaluation phase, we sample the mixture of Bernoulli distributions.

\section{Experiments}
In this section, we elaborate on the datasets and the metrics we have used to validate the efficacy of our proposed model for class-conditional graph generation.



\subsection{Datasets}
\textbf{PROTEINS:} This dataset contains 1113 graphs, having from 100 to 620 nodes. Each graph represents a protein structure, with amino acids as their nodes, and an edge connects two nodes if they are less than 6 Angstroms apart. These graphs are categorized into two enzymes or non-enzymes classes.\\
\textbf{NCI1}: NCI1 is a cheminformatics dataset including 4110 graphs representing chemical compounds classified as positive or negative to cell lung cancer. The compound's molecule atoms form the graph's nodes, and their bonds denote the edges. Their number of nodes ranges from 8 to 111. 

\subsection{Experimental Setup}
We have used three GraphSAGE convolutional layers for our classifier model, with 32 output channels, followed by a two-layer MLP with ReLU non-linearity. The final output of the classifier is equal to the number of classes in each dataset. Then, in CCGG, node representation dimensions are set to 512 for the NCI1 and 384 for the PROTEINS. In each case, a subvector of 256 represents the class data. Recommended by the GRAN paper and confirmed by our results, we have set the number of Bernoulli components to 20 and used seven layers of GNNs. For our node label predictor, we used a three-layer MLP with ReLU non-linearity. Its input is the node representation, where the hidden dimensions of middle layers are set to 256. Also, the output's size equals the number of node labels in each dataset, which is 3 and 37 for the PROTEINS and NCI1 datasets, respectively. Also, we have set $\gamma$'s value to 0.8 for our loss function's calculation. Moreover, we exploited the Adam optimizer for training different parts of our model.


\subsection{Compared Method}
The problem of class-conditional graph generation is a relatively unexplored one. Therefore, there are not many baseline models available for comparison. We test the performance of CCGG against CONDGEN \cite{yang2019conditional}, a formerly proposed method, with state-of-the-art results in conditional graph generation, using Generative Adversarial Nets (GAN). However, since CONDGEN doesn't directly address the class-conditional problem, we have set its condition to the class label for models' equivalence. We have used the same experimental setup during the train and test steps to perform a fair comparison with the baseline.


\subsection{Performance Metrics}
For evaluating the generated graphs, we have used some statistics-based metrics. Our statistics-based metrics are LCC (Largest Connected Component), TC (Triangle Count), Mean D (Mean Degree of Nodes), GINI (Gini index) of the degree distribution \cite{GOSWAMI201816}, and CPL (Characteristic Path Length). Moreover, we have calculated the accuracy and the Area Under the Curve (AUC) for evaluating the classification task of CCGG (Accuracy is calculated separately for each class, as the percentage of total correct classifications of the given class divided by the total number of the generated instances).


\section{Results and Discussion}

In this section, we confirm the efficacy of CCGG by reporting our results on real-world datasets. We have also evaluated CCGG's outputs by two other graph classifiers (in addition to the primary classifier GraphSAGE, which is also utilized in the components of the proposed model), namely, DGCNN \cite{PHAN2018533}, and DiffPool \cite{DBLP:journals/corr/abs-1806-08804}, to support its validity. Note that these two classifiers were not used during the training phase. 

According to the statistics-based metrics shown in Table \ref{gini}, CCGG outperforms the baseline results. Furthermore, we have reported the accuracy of classification in each class of the datasets, as well as the Area Under the Curve (AUC), in the Tables \ref{accuracytable} and \ref{auctable}, respectively. Our model has achieved better results than CONDGEN in both datasets, with few exceptions. Utilizing the class vectors in the model's input and using the node label predictor facilitates the class-conditional generation task. The quality of the generated graphs are also preserved and, in most cases, improved (as confirmed by Table \ref{gini}). This is because the model is trained to minimize the CCGG's loss function in Eq. (\ref{eq:total_loss}), which maintains the balance between each task.

As mentioned before, CCGG’s results were also experimented with using two other classifiers, which were not used during the training phase. One of them is DiffPool, a differentiable graph pooling module that can be used to generate hierarchical representations of graphs by adapting to various GNN architectures, and the other one is DGCNN, an end-to-end deep learning architecture for graph classification, which directly accepts graphs as inputs, without the need of any preprocessing. 

However, despite the superiority of CCGG’s results in most cases, there are still a few places where the baseline outperforms us. We can find the reason behind this by examining the classifiers themselves. The study performed in \cite{errica2019fair} provides a fair comparison of graph classifiers, which shows that the graph classifiers do not guarantee perfect performance. More precisely, according to \cite{errica2019fair}, their classification accuracy never exceeds 80 percent on NCI1 and PROTEINS datasets. Therefore, it brings an intrinsic and unavoidable error in our training phase. We compute the part of our training loss concerning graph classification. We use the imperfect graph classifiers to classify the generated samples in the testing phase. Considering all these, the CCGG model outperforms the baseline, with only a few exceptions. These exceptions mainly occur when DGCNN is employed to classify the generated samples in the test stage, while the model parameters are trained using the GraphSAGE. It is also worth mentioning that according to \cite{errica2019fair}, DGCNN has a relatively lower classification accuracy compared to Diffpool, the other classifier used in the test phase. Our obtained results also show that the performance computed by the DGCNN classifier (as the evaluator) is generally lower both for CCGG and CONDGEN (compared to GraphSAGE and DiffPool), which may be due to the lower capability of the DGCNN classifier itself.





\section{Conclusion}

This paper introduced CCGG, an autoregressive model for solving class-conditioned graph generation problems based on a previously proposed generative model. We used two real-world datasets and achieved state-of-the-art performance on them. Our future direction will extend the model's scalability in generating larger graphs. We hope this work will inspire further research on new applications of class-conditional graph generation.

\bibliographystyle{ACM-Reference-Format}
\bibliography{main}










\end{document}